\documentclass{article}

\usepackage{arxiv}

\usepackage[utf8]{inputenc} 
\usepackage[T1]{fontenc}    
\usepackage{hyperref}       
\usepackage{url}            
\usepackage{booktabs}       
\usepackage{amsfonts}       
\usepackage{nicefrac}       
\usepackage{microtype}      
\usepackage{lipsum}
\usepackage{caption}
\usepackage{subcaption}
\usepackage{graphicx}
\usepackage{hyperref}
\usepackage{multirow}
\usepackage{multicol}

\title{Rice grain disease identification using dual phase convolutional neural network based system aimed at small dataset}

\author{
  Tashin Ahmed\thanks{corresponding author: tashinahmed@aol.com} 
  \And
 Chowdhury Rafeed Rahman 
   \And
 Md. Faysal Mahmud Abid 
}

\begin{document}
\maketitle

\begin{abstract}
Although Convolutional neural networks (CNNs) are widely used for plant disease detection, they require a large number of training samples while dealing with wide variety of heterogeneous background. In this paper, a CNN based dual phase method has been proposed which can work effectively on small rice grain disease dataset with heterogeneity. At the first phase, Faster RCNN method is applied for cropping out the significant portion (rice grain) from an image. This initial phase results in a secondary dataset of rice grains devoid of heterogeneous background. Disease classification is performed on such derived and simplified samples using CNN architecture. Comparison of the dual phase approach with straight forward application of CNN on the small grain dataset shows the effectiveness of the proposed method which provides a 5 fold cross validation accuracy of 88.92\%.
\end{abstract}


\section{Introduction}
As rice grain diseases occur at the very last moment ahead of harvesting, it does major damage to the cultivation process. The average loss of rice due to grain discolouration \cite{baite2019disease} was 18.9\% in India. Yield losses caused by False Smut (FS) \cite{atia2004rice} ranged from 1.01\% to 10.91\% in Egypt. 75\% yield loss of grain occurred in India in 1950, while in the Philippines \cite{ou1985rice} more than 50\% yield loss was recorded. Rice yield loss is a direct consequence of Neck Blast (NB) disease, since this disease results in poor panicles. A big reason behind Neck Blast \cite{khan2014neck} is an extreme phase of the Blast and grain disease. In Bangladesh False Smut was one of the most destructive rice grain disease \cite{nessa2017rice} from year 2000 to 2017. 

Collecting field level data on agronomy is a challenging task in the context of poor and developing countries. The challenges include lack of equipment and specialists. Farmers of such areas are ignorant of technology use which makes it quite difficult to collect crop disease related data efficiently using smart devices via the farmers. Hence, scarcity of plant disease oriented data is a common challenge while automating disease detection in such areas.

\begin{figure}
\captionsetup[subfigure]{labelformat=empty}
 \centering
 \begin{subfigure}{1.0\textwidth}
  \centering
  \includegraphics[width=1.0\linewidth]{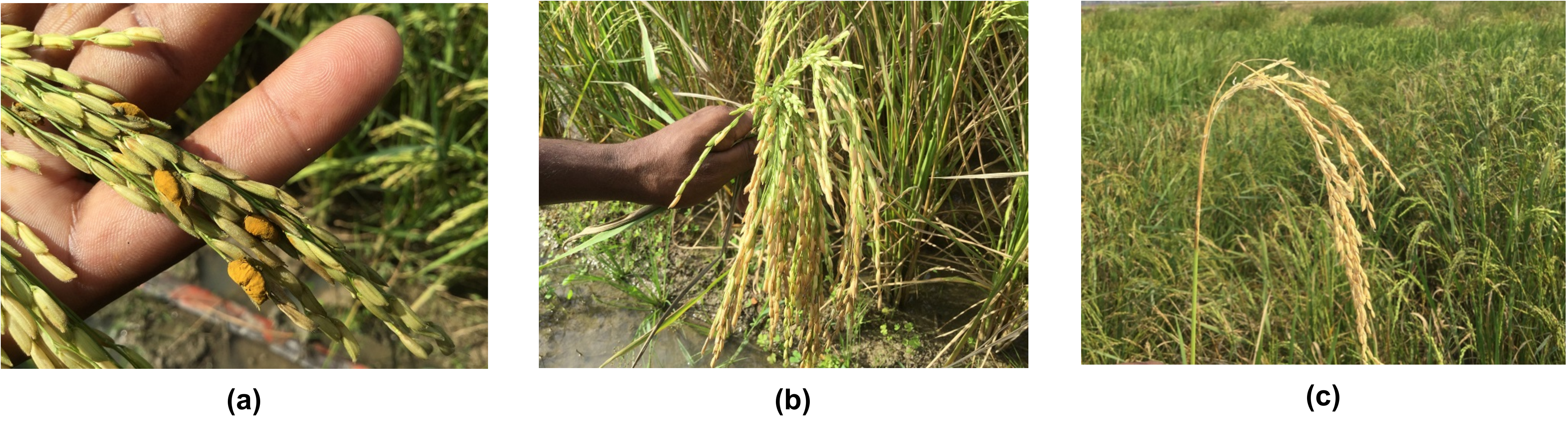}
 \end{subfigure}
\caption{Sample image of every grain classes. From left: (a) False Smut, (b) Healthy and (c) Neck Blast. Images clearly show the complexity of the dataset where foreground and background are quite identical.}\label{fig:three_images.pdf}
\end{figure}

Many researches have been undertaken with a view to automating plant disease detection utilizing different techniques of machine learning and image processing. \cite{pugoy2011automated} through system with the ability to identify areas which contain abnormalities. They applied a threshold based clustering algorithm for this task. A framework have been created \cite{sethy2017detection} for the detection of defected diseased leaf using K-Means clustering based segmentation. They claimed that their approach was able to detect the healthy leaf area and defected diseased area accurately. A genetic algorithm has been developed \cite{chung2016detecting} which was used for selecting essential traits and optimal model parameters for the SVM classifiers for Bakanae \textit{gibberella fujikuroi} disease. A technique to classify the diseases based on percentage of RGB value of the affected portion was proposed \cite{islam2018faster} utilizing image processing. A similar technique using multi-level colour image thresholding was proposed \cite{bakar2018rice} for RLB disease detection. Deep learning based object classification and segmentation has become the state-of-the-art for automatic plant disease detection. Neural network was employed \cite{babu2007leaves} for leaf disease recognition while a self organizing map neural network (SOM-NN) was used \cite{phadikar2008rice} to classify rice disease images. Researchers also experimented with AlexNet \cite{atole2018multiclass} to distinguish among 3 classes of rice disease using a small dataset containing 227 images. A similar research for classifying 10 classes of rice disease on a 500 image dataset was undertaken \cite{lu2017identification} using a handmade deep CNN architecture. Furthermore, the benefit of using pre-trained model of AlexNet and GoogleNet \cite{brahimi2017deep} has been demonstrated when the training data is not large. Their dataset consisted of 9 diseases of tomatoes. 
A detailed comparative analysis of different state-of-the-art CNN baseline and finely tuned architecture \cite{rahman2018identification} on eight classes of rice disease and pest also conveys a huge potential. It demonstrates two-stage training approach for memory efficient small CNN architectures. 
Besides works on rice disease some experimental procedure on other agricultural elements are, \cite{ferentinos2018deep} specialized deep learning models based on specific CNN architectures for identification of plant leaf diseases with a dataset containing 58 classes from 25 different plants have been developed. 
Transfer learning approach using GoogleNet \cite{barbedo2018impact} on a dataset containing 87848 images of 56 diseases infecting 12 plants. Extraction of disease region from leaf image through different segmentation \cite{shah2016survey} techniques was a driving step. Some of the image segmentation algorithms were compared \cite{devi2014analysis} in order to segment the diseased portion of rice leaves. 

Though the above mentioned researches have a significant contribution to the automation of disease detection, none of the works addressed the problem of scarcity of data which limits the performance of CNN based architectures. Most of the researches focused on image augmentation techniques to tackle the dataset size issue. But applying different geometric augmentations on small size images \cite{liu2016overfitting,shorten2019survey} result in nearly the same type of image production which has drawbacks in terms of neural network training. Production of similar images through augmentation \cite{cogswell2015reducing} can cause overfitting as well.

The first phase of the proposed method deals with a learning oriented segmentation based architecture. This architecture helps in detecting the significant grain portion of a given image that has a heterogeneous background, which is an easier task compared to disease localization. The detected grain portions cropped from the original image are used as separate simplified images. In the second phase, these simplistic grain images are used in order to detect grain disease using fine tuned CNN architecture. Because of the simplicity of the tasks assigned in the two phases, our proposed method performs well in spite of having only 200 images of three classes. To prove the proposed approach is satisfactory for a small dataset, counter experimentations on straightforward CNN and Faster RCNN were also demonstrated at Section \ref{results and discussion}.
\newline

\begin{figure}
\captionsetup[subfigure]{labelformat=empty}
 \centering
 \begin{subfigure}{1.0\textwidth}
  \centering
  \includegraphics[width=1.0\linewidth]{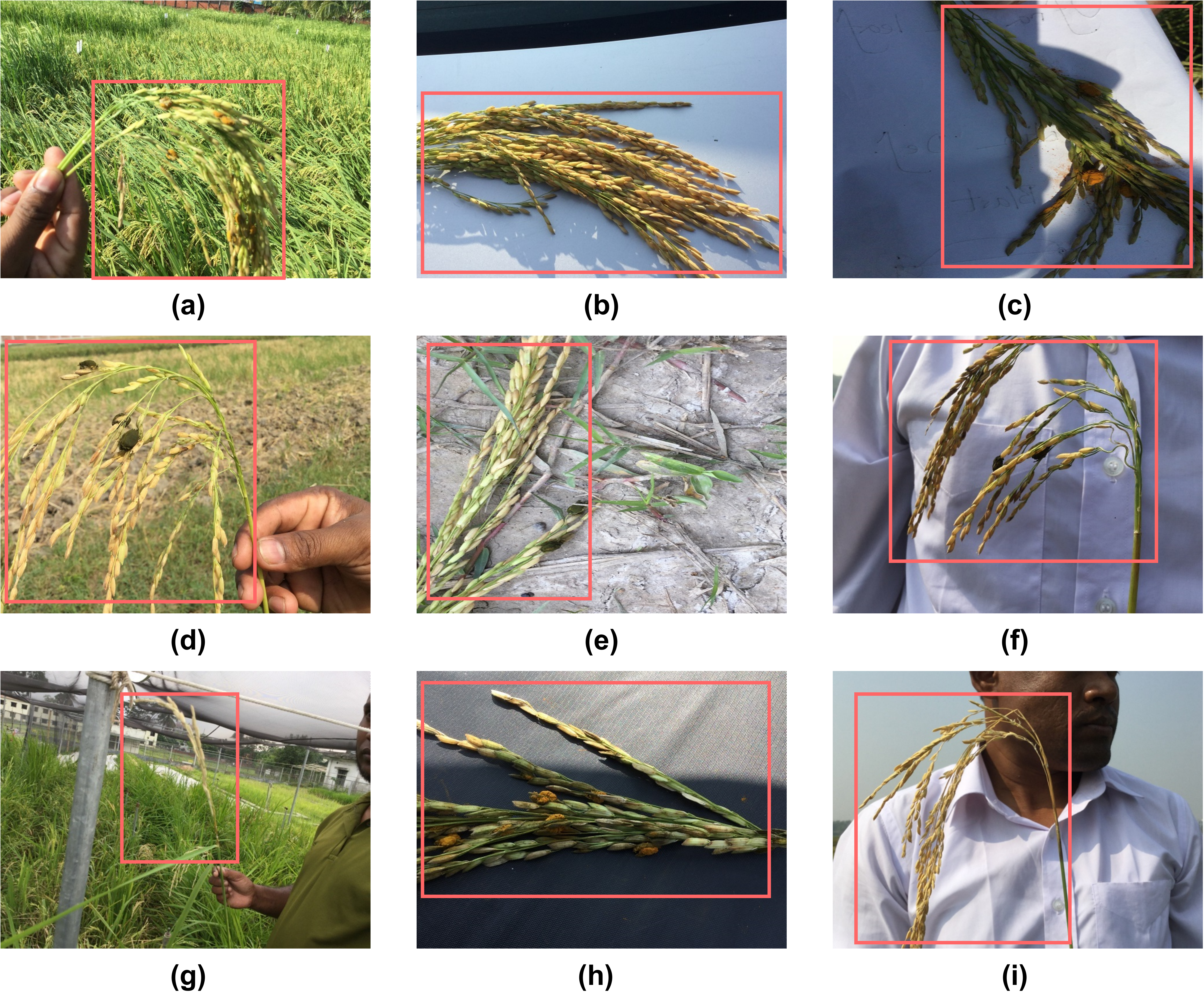}
 \end{subfigure}
\caption{Sample of data heterogeneity amongst the dataset which demonstrates different parameters like unique backgrounds, lights, contrast, distance have been taken into account while data was gathered. Red rectangular boxes show how the annotation process has been performed. (a, d, f): multiclass images consists of Neck Blast and False Smut, (b): healthy, (c, e, h): False Smut, (g, i): Neck Blast.}\label{hetero}
\end{figure}

\section{Our Dataset}\label{dcapp}
Our balanced dataset of 200 images consists of three classes - False Smut, Neck Blast and healthy grain class as shown in Table \ref{primary dataset}. 

\begin{table}
\centering
\begin{tabular}{lll}
\hline
Class & \begin{tabular}[c]{@{}l@{}}Image\\ Count\end{tabular} & \begin{tabular}[c]{@{}l@{}}Image\\ Percentage\end{tabular} \\ \hline
False Smut & 75          & 37.5          \\
Neck Blast & 63          & 31.5          \\
Healthy    & 62          & 31.0         \\ \hline
\end{tabular}
\caption{Image count and percentage(\%) of each class of the primary dataset.}
\label{primary dataset}
\end{table}

A sample image from each class has been shown in Figure. \ref{fig:three_images.pdf}. Neck Blast is generally caused by the fungus known as \textit{Magnaporthe oryzae}. It causes plants to develop very few or no grains at all. Infected nodes result \cite{wilson2009under} in panicle break down. False Smut is caused by the fungus called \textit{Ustilaginoidea virens}. It results in lower grain weight and reduction \cite{koiso1994ustiloxins} of seed germination.

Data have been collected and annotated from two separate sources for this experiment - field data supervised by officials from Bangladesh Rice Research Institute (BRRI) and image data from a previously expermineted \cite{rahman2018identification} repository. As Boro species have the maximum threat to be affected with False Smut and Neck Blast, Boro rice plant has been chosen \cite{miah1985survey} for experimental data collection. Parameters like light, distance and uniqueness have been taken into consideration while capturing the photographs. The main parameter which was taken into account was heterogeneity of the background. Some sample images of hetergenous background of the dataset have been presented in Figure. \ref{hetero}.

\begin{figure}
\captionsetup[subfigure]{labelformat=empty}
 \centering
 \begin{subfigure}{1.0\textwidth}
  \centering
  \includegraphics[width=1.0\linewidth]{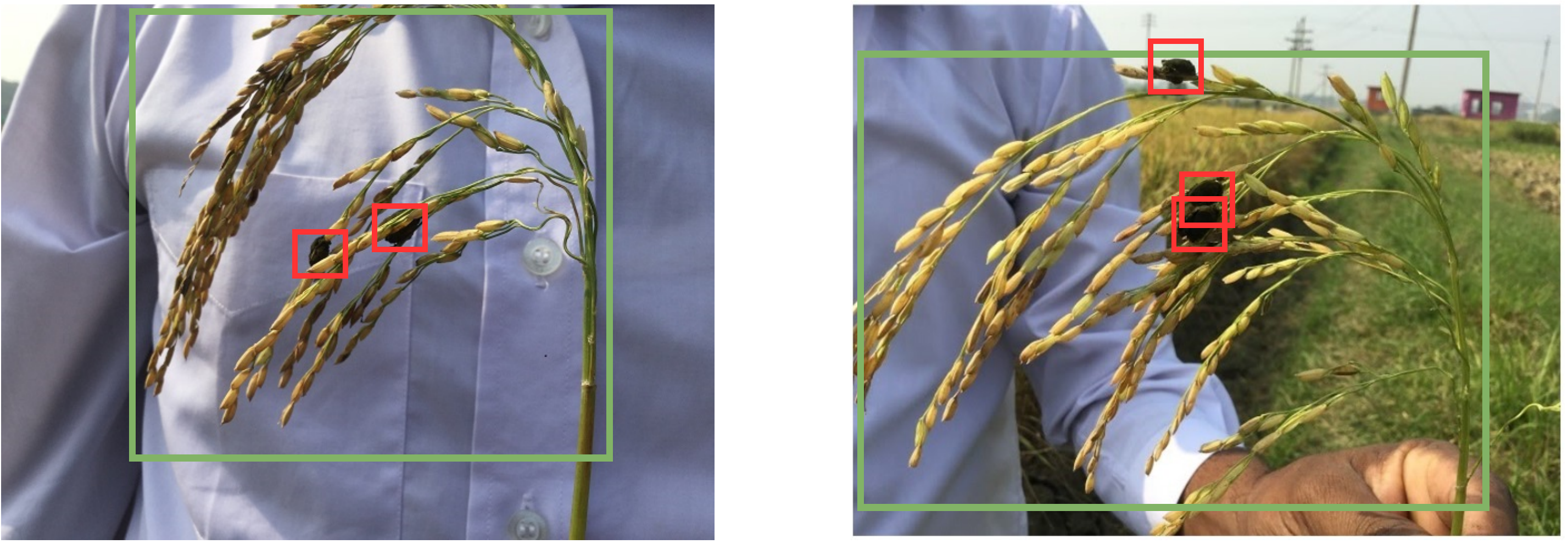}
 \end{subfigure}
\caption{Sample multiclass data; Labeled green for Neck Blast and red for False Smut class}\label{fig:multiclass.pdf}
\end{figure}

To make the dataset more challenging multiclass images also have been taken into account. Sample images of multiclass data which consists both Neck Blast and False Smut class have been presented in Figure. \ref{fig:multiclass.pdf}. If there is a multiclass data it had been labelled as Neck Blast for the training phase of counter experiments (explained in Section \ref{results and discussion}) as False Smut already surpassed by quantity than Neck Blast. Also, the dataset was split into 80:20 in terms of train and validation set. Augmentation techniques were not applied as the main goal of this experiment is to use small and natural scene image data. Additionally, there are other factors that can spoil the experiment, such as illumination, symptom severity, maturity of the plant and diseases. A large versatile dataset can attend on that occasion which can be achieved in the future. The dataset has been kept to three classes for the early stages of the investigation. 
Also, it is quite challenging and burdensome to collect different rice disease image dataset throughout the year as different diseases occur at different time. So, at this early stage of the investigation three classes is competent to deliver a sufficient result. Supplementary public data related to the paper can be found at \url{https://cutt.ly/RvgxoDi}.

\section{Materials and Methods}\label{materials and methods}

\subsection{Experimental Setup}\label{es}
This subsection explained about the experimental setup which includes hardware used in this experiment and explains five basenets applied. This subsection have also discussed about different hyperparameter optimization and their consequence in this experiment.

\subsubsection{Hardware}\label{hardware}
For the training environment, assistance has been taken from two different sources.

\begin{itemize}
    \item Royal Melbourne Institute of Technology \href{https://www.rmit.edu.au/}{(RMIT)} provides GPU for international research enthusiasts and they provided a Red Hat Enterprise Linux Server along with the processor Intel Xeon E5-2690 CPU, clock speed of 2.60 GHz. It has 56 CPUs with two threads per core, 503 GB of RAM. Each user can use up to 1 petabyte of storage. There are also two 16 GB NVIDIA Tesla P100-PCIE GPUs available. First phase was completed through this server. 
    \item  Google Colab (Tesla K80 GPU, 12GB RAM) and Kaggle kernel (Tesla P100 GPU) have been used for counter experimentation. 
\end{itemize}

\subsubsection{Utilized CNN Models}\label{our models}
\begin{figure}
\captionsetup[subfigure]{labelformat=empty}
 \centering
 \begin{subfigure}{1.0\textwidth}
  \centering
  \includegraphics[width=1.0\linewidth]{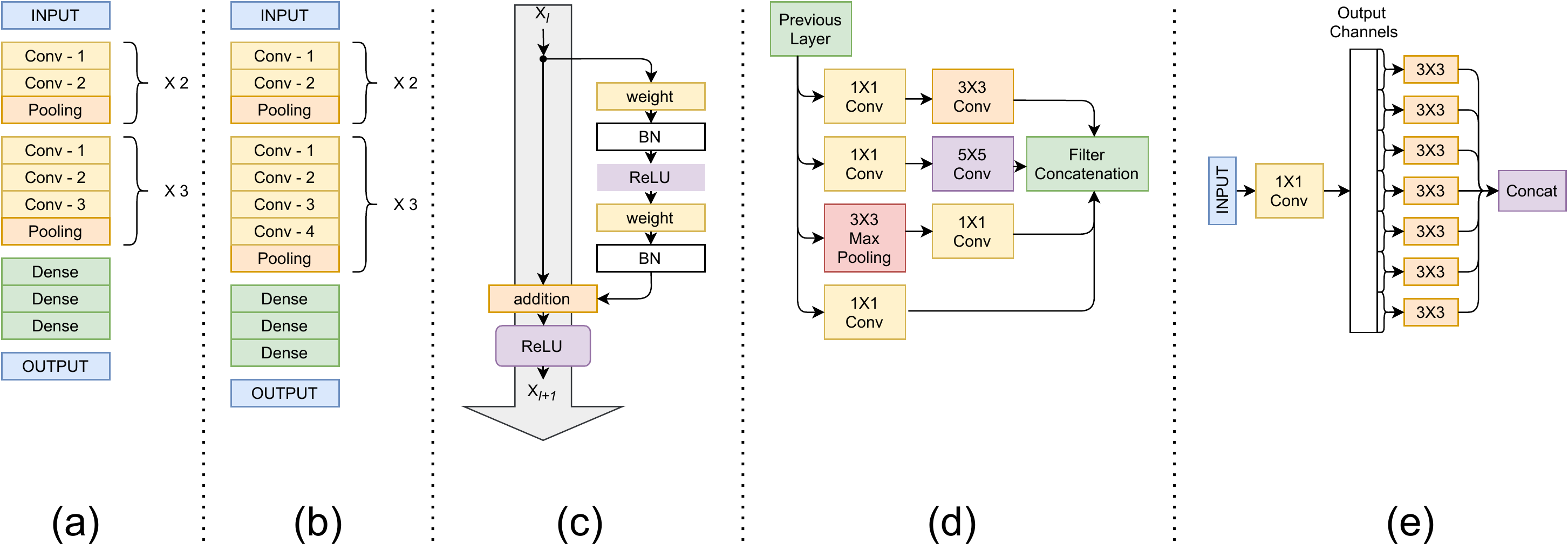}
 \end{subfigure}
\caption{(a): all 16 blocks of VGG16, (b): all 19 blocks of VGG19, (c): Resdiual module of ResNet, (d): Inception module is to act as a multi-level feature extractor in InceptionV3, (e): Extreme module of the Inception module which is utlized in Xception.}\label{fig:cnn.pdf}
\end{figure}

Experiments have been performed using five state-of-the-art CNN architectures which are described as follows. 
Figure. \ref{fig:cnn.pdf} shows architectures and key blocks of the applied CNN architectures.

\begin{description}
\item[VGG16] is a sequential architecture \cite{simonyan2014very} consisting of 16 convolutional layers. Kernel size in all convolution layers is three. 

\item[VGG19] has three extra convolutional layers \cite{simonyan2014very} and the rest is the same as VGG16.

\item[ResNet50] belongs to the family of residual neural networks. It is a deep CNN architecture \cite{he2016deep} with skip connections and batch normalization. The skip connections help in eliminating the gradient vanishing problem. 

\item[InceptionV3] is a CNN architecture \cite{szegedy2016rethinking} with parallel convolution branching. Some of the branches have filter size as large as $7\times7$.

\item[Xception] takes the principles of Inception to an extreme. Instead of partitioning the input data into several chunks, it maps the spatial correlations \cite{chollet2017xception} for each output channel separately and performs $1\times1$ depthwise convolution. 
\end{description}

\subsubsection{Optimized Hyperparameters}\label{tuned hyperparemeters}
Hyperparameters of Faster RCNN have been presented in Table \ref{table_hyperparam}. 

\begin{table}
\centering
\begin{tabular}{ll}
\hline
\textbf{Hyperparameter}          & \textbf{Optimized Value}       \\ \hline
Anchor Box Count                 &9, \textbf{16}              \\ \hline
Anchor Box Size (pixels)         &[\textbf{32, 64, 128, 256}], [128, 256, 512]  \\\hline
Anchor Box Ratios                &[(1,1), (2,1), (1,2)], [\textbf{(1,1), (\begin{math}\mathbf{\frac{1}{\sqrt2},\frac{2}{\sqrt2}}\end{math}),
(\begin{math}\mathbf{\frac{2}{\sqrt2},\frac{1}{\sqrt2}}\end{math}), 
(2,2)}]  \\\hline

RPN Threshold                   & 0.3 - 0.7, \textbf{0.4 - 0.8}         \\\hline
Proposal Selection      & \textbf{200}, 2000               \\\hline

Overlap Threshold &\textgreater{}\textbf{0.8}, \textgreater{}0.9 \\\hline

Learning Rate &0.001, \textbf{0.0001}, 0.00001\\\hline
Optimizers              & \textbf{Adam}, SGD              \\ \hline
\end{tabular}
\caption{Experimented hyperparameters. Bold values were selcted for the prime experiment.}
\label{table_hyperparam}
\end{table}

\begin{description}

\item[Anchor Box Hyperparameters:] Anchor boxes are a set of bounding boxes defined through different scales and aspect ratios. They mark the probable regions of interest of different shapes and sizes. The total number of probable anchor boxes per pixel of a convolutional feature map is $P_{n} \times R_{n}$, where $P_{n}$ and $R_{n}$ denote the number of anchor box size variations and ratio variations respectively.  

\item[Region Proposal Network (RPN) Hyperparameters:] RPN layer utilizes the convolutional feature map of the original image to propose regions of interest that are identifiable within the original image. The proposals are made in line with the anchor boxes. For each anchor box, RPN predicts if it is an object of interest or not and changes the size of the anchor box to better fit the object. RPN threshold of 0.4 - 0.8 means that any proposed region which has IoU (Intersection Over Union) less than 0.4 with ground truth object is considered a wrong guess whereas any proposed region which has IoU greater than 0.8 with ground truth object is considered correct. This notion is used for training the RPN layer. 

\item[Proposal Selection:] Proposal selection threshold of 200 means that top (according to probability) 200 region proposals from RPN layer will pass on to the next layers for further processing. 

\item[Overlap Threshold:] During non-max suppression, overlapping object proposals are excluded if the IoU is above a certain threshold. If their overlap is greater than the threshold, only the proposal with the highest probability is kept and the procedure continues until there are no more boxes with sufficient overlap.

\item[Learning Rate:] It is used for controlling the speed of model parameter update.  

\item[Optimizer:] Optimizer is an algorithm for updating model parameter weights after training on each batch of samples. Weight updating process varies with the choice of optimizer.   
\end{description}

\subsection{Proposed Dual Phase Approach}\label{PDPA}
\begin{figure}
\captionsetup[subfigure]{labelformat=empty}
 \centering
 \begin{subfigure}{1.0\textwidth}
  \centering
  \includegraphics[width=0.9\linewidth]{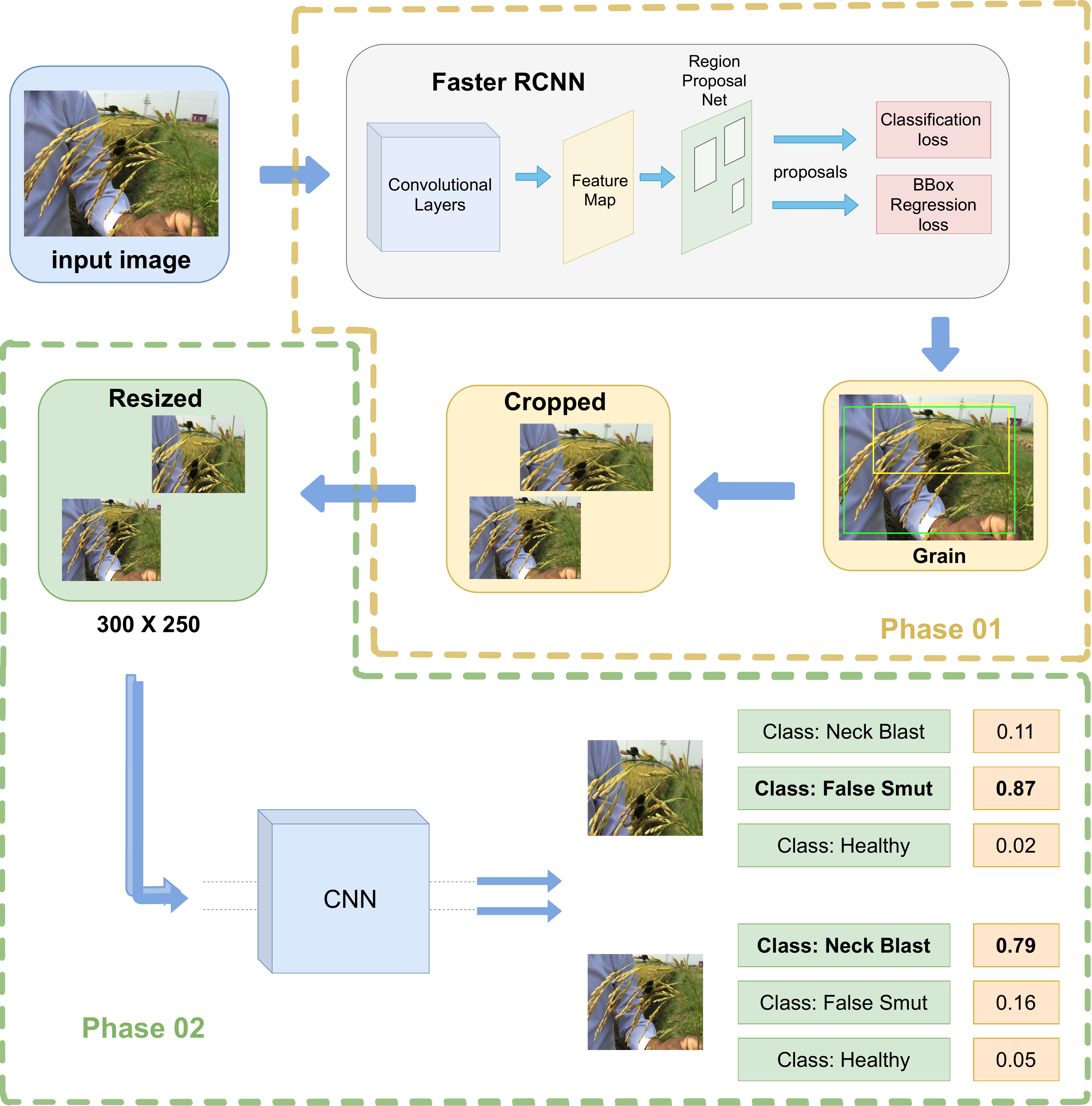}
 \end{subfigure}
\caption{Proposed dual phase approach; Phase one for detection of the significant portion and phase two for classification; A multiclass data have been presented as an example to demonstrate the classification strategy. }\label{fig: FasterRCNN.pdf}
\end{figure}

In this research, a dual phase approach has been introduced in order to learn effectively from small dataset containing images with a lot of heterogeneity in the background. The approach overview has been provided in Figure. \ref{fig: FasterRCNN.pdf}. In the first phase, the original image is taken, reshaped to a fixed size and then passed through segmentation oriented Faster RCNN architecture. At most two best regions have been selected from the first phase. After obtaining the significant grain portions from an image, those regions are cropped and resized to a fixed size. These images look simple because of the absence of heterogeneous background. CNN architecture is trained on this simplified dataset to detect disease. The learning process has been showed to be effective through experiments. 

\subsubsection{Segmenting Grain Portion}\label{SGP}

This is the first phase of our approach. Segmentation algorithms based on CNN architecture as a backbone requires image to be of fixed size. Input images have been resized to $640\times$480 before feeding them to Faster RCNN. The consecutive stages of the network through which this resized image passes through have been described as follows.

\begin{description}

\item[Convolutional Neural Network (CNN):]
In order to avoid sliding a window in each spatial position of the original image, CNN architecture is used in order to learn and extract feature map from the image which represents the image effectively. The spatial dimension of such feature map decreases whereas the channel number increases. For the dataset used in this research, VGG16 architecture has proven to be the most effective. Hence, VGG16 has been used as the backbone CNN architecture which transforms the original image into $20\times15\times512$ dimension.

\item[Region Proposal Network (RPN):]
The extracted feature map is passed through RPN layer. For each pixel of the feature map of spatial size 20$\times$15, there are 16 possible bounding boxes (4 different aspect ratios and 4 different sizes mentioned in bold letter in Table \ref{table_hyperparam}). So, that makes total 16$\times$20$\times$15 = 4800 possible bounding boxes, RPN is a two branch Convolution layer which provides two scores (branch one) and four coordinate adjustments (branch two) for each of the 4800 boxes. The two scores correspond to the probability of being an object and a non-object. Only those boxes which have a high object probability are taken into account. Non-max suppression (NMS) is used in order to eliminate overlapping object bounding boxes and to keep only the high probability unique boxes. The threshold of this overlap in this case is 0.8 IoU. From this probable object proposals, top 200 proposals according to object probability are passed to the next layers.

\item[ROI Pooling:]
Each of the 200 selected object proposals correspond to some region in the CNN feature map. For passing each of these regions on to the dense layers of the architecture, each of the regions need to be of fixed size. ROI pooling layer takes each region and turns them into 7$\times$7$\times$512 using bilinear interpolation and max pooling.

\item[RCNN Layer:]
RCNN layer consists of fully connected dense layers. Each of the 7$\times$7$\times$512 size feature maps are flattened and passed through these fully connected layers. The final layer has two branches. Branch one predicts if the input feature map is background class or significant grain portion. Branch two provides four regression values denoting the adjustment of the bounding box to better fit the grain portion. For each feature map, if the probability of being a grain is over 0.6, only then is the feature map considered as a probable grain portion and the adjusted coordinates are mapped to the original image in order to get the localized grain portion. The overlapping boxes are eliminated using NMS. The remaining bounding box regions are the significant grain portions.

\item[Loss Function:]
The trainable layers of Faster RCNN architecture are: CNN backbone, RPN layer and RCNN layer. A loss function is needed in order to train these layers in an end to manner which is as follows. 

$$L({p_{i}},{t_{i}}) = \frac{1}{N_{cls}}\sum\limits_{i} L_{cls}(p_{i},p_{i}^*) + 
\lambda\frac{1}{N_{reg}} \sum\limits_{i} p_{i}^*L_{reg}(t_{i},t_{i}^*) $$

The first term of this loss function defines the classification loss over two classes which describe whether predicted bounding box $i$ is an object or not.
The second term defines the regression loss of the bounding box when there is a ground truth object having significant overlap with the box.
Here, $p_{i}$ and $t_{i}$ denote predicted object probability of bounding box $i$ and predicted four coordinates of that box respectively while $p_{i}^{*}$ and $t_{i}^{*}$ denote the same for the ground truth bounding box which has enough overlap with predicted bounding box $i$. $N_{cls}$ is the batch size (256 in this case) and $N_{reg}$ is the total number of bounding boxes having enough overlap with ground truth object. Both these terms work as normalization factor. $L_{cls}$ and $L_{reg}$ are log loss (for classification) and regularized loss (for regression) function respectively.  
\end{description}

\subsubsection{Disease Detection from Segmented Grain}\label{DDFG}
Figure. \ref{fig: FasterRCNN.pdf} shows Faster RCNN architecture drawing bounding boxes on two significant grain portions. These portions are cropped and resized to a fixed size (300$\times$250 in this case) in order to pass each of them through a CNN architecture. Thus two images have been created from single image of the primary dataset.
The same process can be executed on each of the images of the primary dataset. Thus a secondary dataset of significant grain portion can be created. Each of these images have to be labeled as one of the three classes in order to train the CNN architecture. The complete dataset including these secondary image counts has been shown in Table \ref{Complete Dataset}. The cropped portions when passed through a trained CNN model have been predicted as False Smut disease and Neck Blast class in Figure. \ref{fig: FasterRCNN.pdf} as it is an example of multiclass data.

\begin{table}[]
\centering
\begin{tabular}{llll}
\hline
\multirow{2}{*}{Class} & \multicolumn{2}{c}{Image Count} & \multirow{2}{*}{\begin{tabular}[c]{@{}l@{}}Image\\ Increment\end{tabular}} \\ \cline{2-3}
           & Primary & Secondary &    \\ \hline
False Smut & 75      & 85        & 10 \\
Neck Blast & 63      & 70        & 7  \\
Healthy    & 62      & 64        & 2  \\ \hline
Total      & 200     & 219       & 19 \\ \hline
\end{tabular}
\caption{Complete Dataset}
\label{Complete Dataset}
\end{table}

\subsection{Evaluation Metric}
All results have been provided in terms of 5 fold cross validation. 
Accuracy metric has been utilized in order to compare dual phase approach against implementation of CNN on original images without any segmentation. Accuracy is a suitable metric for balanced dataset.  

$$Accuracy = \frac{TP}{TP+FP+TN+FN}\footnote{TP: True Positive, FP: False Positive, TN: True Negative, FN: False Negative}$$

Segmenting the grain portion is the goal of the first phase of the dual phase approach. For evaluating the performance of this phase, mAP (mean average precision) score has been used. Precision, recall and IoU (Intersection over Union) are required to calculate mAP score.

$$Precision = \frac{TP}{TP+FP}$$
$$Recall = \frac{TP}{TP+FN}$$
$$IoU = \frac{AOI}{AOU}\footnote{AOI: Area of intersection, AOU: Area of union (with respect to ground truth bounding box)}$$

If a predicted box IoU is greater than a certain predefined threshold, it is considered as TP. Otherwise, it is considered as FP. $(TP+FN)$ is actually the total number of ground truth bounding boxes. Average precision (AP) is calculated from the area under the precision-recall curve. If there are N classes, then mAP is the average AP of all these classes. In this research, there is only one class of object in phase one, that is the significant grain portion class. So, here AP and mAP are the same. 

As the proposed method has two stages: segmentation and classification, failure in proper segmentation can leads to classification failure. In this work, two stages are created as an intact pipeline so that outcome of the first stage directed to the second stage as input. Detail about the procedure mentioned in Subsection \ref{prime_experiment}.

\section{Results and Discussion}\label{results and discussion}
As mentioned earlier, the proposed dual-phase approach has been performed in two steps. First, segmentation of the grain parts and lastly the classification of the segmented parts. This experiment has been mentioned as the prime experiment throughout the paper. To verify the performance of the prime experiment, different CNN architectures and Faster RCNN has been employed separately. This part of the experiment has been mentioned as the counter experiment. Individual counter experiments have been performed to analyze their performance with the respective phase of the prime experiment.

\begin{table}
\centering
\begin{tabular}{llll}
\hline
\begin{tabular}[c]{@{}l@{}}Transfer\\ learning\\ Approach\end{tabular} &
  \begin{tabular}[c]{@{}l@{}}CNN\\ Architecture\end{tabular} &
  \begin{tabular}[c]{@{}l@{}}Validation\\ Loss\end{tabular} &
  \begin{tabular}[c]{@{}l@{}}Validation\\ Accuracy\\ (\%)\end{tabular} \\ \hline
\multirow{5}{*}{\begin{tabular}[c]{@{}c@{}}Freezed\\Layer\end{tabular}}           & \textbf{VGG16}       & \textbf{2.08}  & $\textbf{63.33} \pm \textbf{2.04}$  \\
                                                                                   & VGG19       & 1.08  & $43.75 \pm 3.43$  \\
                                                                                   & Xception    & 2.34  & $31.25 \pm  4.04$ \\
                                                                                   & InceptionV3 & 9.23  & $37.50 \pm 3.89$  \\
                                                                                   & ResNet50    & 4.47  & $31.25 \pm 3.27$  \\ \hline
\multirow{5}{*}{\begin{tabular}[c]{@{}c@{}}Fine\\Tuned\end{tabular}}             & \textbf{VGG16}       & \textbf{2.71}  & $\textbf{67.79} \pm \textbf{3.24}$  \\
                                                                                   & VGG19       & 1.77  & $55.04 \pm 3.00$  \\
                                                                                   & Xception    & 6.29 & $43.76 \pm 1.88$  \\
                                                                                   & InceptionV3 & 7.42 & $41.73 \pm 3.66$  \\
                                                                                   & ResNet50    & 2.47  & $38.20 \pm 1.34$  \\ \hline
\multirow{5}{*}{\begin{tabular}[c]{@{}c@{}}Fine Tuned\\+\\Dropout\end{tabular}} & \textbf{VGG16}       & \textbf{3.47}  & $\textbf{69.43} \pm \textbf{3.41}$  \\
                                                                                   & VGG19       & 3.11  & $57.18 \pm 2.64$  \\
                                                                                   & Xception    & 5.72 & $47.17 \pm 2.11$  \\
                                                                                   & InceptionV3 & 4.12 & $48.22 \pm 3.14$  \\
                                                                                   & ResNet50    & 2.81  & $42.31 \pm 1.32$ \\ \hline
\end{tabular}
\caption{Counter Experiment 01: CNN}
\label{ce1}
\end{table}

\subsection{Counter Experiments}
Two counter experiments have been performed named, counter experiment 01 and 02. Counter experiment 01 is based on various CNN architectures where the goal is to obtain the classification outcome from the primary dataset. Counter experiment 02 is based on Faster RCNN underlying three selected CNN architectures which has been applied for both classification and detection of the three classes.

\subsubsection{Counter Experiment 01: CNN}\label{counter01}
This experiment has been conducted applying five different CNN architectures which were mentioned earlier in Subsubsection \ref{our models}. Three transfer learning approaches have been followed (which are freezed layer, fine tuning and fine tuning + dropout) in this part utilizing imagenet pretrained models. At first, freezing layer approach has been performed which is also known as a default transfer learning method. VGG16 outperformed other CNN architectures with a validation accuracy of $63.33 \pm 2.04$ mentioned in Table \ref{ce1}. 
Then, finely-tuned approach has been applied which shows improvement in validation accuracy of $67.79 \pm 3.24$ for VGG16. Finally, dropout has been applied inside the CNN architectures which results in a significant improvement of $69.43 \pm 3.41$ for VGG16. Fine-tuning and fine-tuning + dropout have been performed several times by experimenting with dropout on various positions inside individual CNNs. Although the standard deviation of the outcome for VGG16 is large which is an indication of low precision. Comparative results for all five architecture have been shown in Table \ref{ce1}.

\subsubsection{Counter Experiment 02: Faster RCNN}\label{counter02}
In this counter experiment 02, Faster RCNN has been applied utilizing three different CNN architectures as the backbone. The goal of this experiment is to test the ability of Faster RCNN for efficient detection and classification of the significant portion (grain).
VGG16 and VGG19 have been chosen because of their performance at counter experiment 01. Additionally, ResNet50 have chosen because of the lower validation loss than Xception and InceptionV3, mentioned in Table \ref{ce1}. CNN models have been applied as pretrained model accumulated from COCO and Imagenet. Different hyperparameter optimizations have been applied to reach the peak outcome for Faster RCNN mentioned in Table \ref{ce2}.

Default settings from the Faster RCNN paper \cite{ren2015faster} for anchor box ratio were (1:1), (2:1), (1:2) and anchor box pixels were 128, 256, 512. Which produces 3$\times$3=9 anchor boxes per pixel. The default RPN threshold of (0.3 - 0.7), overlap threshold 0.8 and default anchor box ratios and pixels, VGG16 (imagenet pretrained model) provided the best mAP score of $71.0 \pm 4.0$. After tuning RPN threshold to (0.4 - 0.8), anchor box ratios to (1:1), (\begin{math}\frac{1}{\sqrt2}:\frac{2}{\sqrt2}\end{math}), (\begin{math}\frac{2}{\sqrt2}:\frac{1}{\sqrt2}\end{math}), (2:2) and pixel sizes to 32, 64, 128, 256, 4$\times$4=16 anchor boxes have been produced which provides better outcome than before. This setting improved the mAP for VGG16 (imagent pretrained model) to $76.32 \pm 2.29$ which is the peak outcome after several optimization.

\begin{table}
\centering
\begin{tabular}{cclllll}
\hline
\begin{tabular}[c]{@{}l@{}}Pretrained\\ Model\end{tabular} &
  \begin{tabular}[c]{@{}l@{}}Anchor \\ Box Ratio\end{tabular} &
  \begin{tabular}[c]{@{}l@{}}Anchor\\ Box Pixels\end{tabular} &
  \begin{tabular}[c]{@{}l@{}}RPN \\ Threshold\end{tabular} &
  \begin{tabular}[c]{@{}l@{}}CNN\\ Architecture\end{tabular} &
  \begin{tabular}[c]{@{}l@{}}Overlap\\ Threshold\end{tabular} &
  \begin{tabular}[c]{@{}l@{}}mAP (\%)\end{tabular} \\ \hline
\multirow{9}{*}{Imagenet} &
  \multirow{3}{*}{\begin{tabular}[c]{@{}c@{}}(1:1), (2:1), \\ (1:2)\end{tabular}} &
  \multirow{3}{*}{\begin{tabular}[c]{@{}c@{}}128,256,\\ 512\end{tabular}} &
  \multirow{3}{*}{0.3 - 0.7} &
  VGG16 &
  \multirow{3}{*}{\textgreater 0.8} &
  $71.0 \pm 4.0$ \\
 &  &  &  & VGG19                     &                  & $47.06 \pm 2.01$  \\
 &  &  &  & ResNet50                  &                  & $67.14 \pm 6.68$  \\ \cline{2-7} 
 &
  \multirow{6}{*}{\begin{tabular}[c]{@{}c@{}}(1:1), \\ (\begin{math}\frac{1}{\sqrt2}:\frac{2}{\sqrt2}\end{math}), \\(\begin{math}\frac{2}{\sqrt2}:\frac{1}{\sqrt2}\end{math}), \\ (2:2)\end{tabular}} &
  \multirow{6}{*}{\begin{tabular}[c]{@{}c@{}}32, 64,\\ 128, 256\end{tabular}} &
  \multirow{6}{*}{0.4 - 0.8} &
  \multirow{2}{*}{\textbf{VGG16}} &
  \textgreater \textbf{0.8} &
  $\textbf{76.32} \pm \textbf{2.29}$ \\
 &  &  &  &                           & \textgreater 0.9 & $63.42 \pm 2.36$  \\ \cline{5-7} 
 &  &  &  & \multirow{2}{*}{VGG19}    & \textgreater 0.8 & $70.08 \pm4.54$   \\
 &  &  &  &                           & \textgreater 0.9 & $70.30 \pm 2.36$  \\ \cline{5-7} 
 &  &  &  & \multirow{2}{*}{ResNet50} & \textgreater 0.9 & $40.02 \pm 3.03$  \\
 &  &  &  &                           & \textgreater 0.8 & $52.36  \pm 5.91$ \\ \hline 
 
 \multirow{9}{*}{COCO} &
  \multirow{3}{*}{\begin{tabular}[c]{@{}c@{}}(1:1), (2:1), \\ (1:2)\end{tabular}} &
  \multirow{3}{*}{\begin{tabular}[c]{@{}c@{}}128,256,\\ 512\end{tabular}} &
  \multirow{3}{*}{0.3 - 0.7} &
  VGG16 &
  \multirow{3}{*}{\textgreater 0.8} &
  $48.32 \pm 4.79$ \\
 &  &  &  & VGG19                     &                  & $32.30 \pm 4.83$  \\
 &  &  &  & ResNet50                  &                  & $46.36 \pm 2.04$  \\ \cline{2-7} 
 &
  \multirow{6}{*}{\begin{tabular}[c]{@{}c@{}}(1:1), \\ (\begin{math}\frac{1}{\sqrt2}:\frac{2}{\sqrt2}\end{math}), \\(\begin{math}\frac{2}{\sqrt2}:\frac{1}{\sqrt2}\end{math}), \\ (2:2)\end{tabular}} &
  \multirow{6}{*}{\begin{tabular}[c]{@{}c@{}}32, 64,\\ 128, 256\end{tabular}} &
  \multirow{6}{*}{0.4 - 0.8} &
  \multirow{2}{*}{\textbf{VGG16}} &
  \textgreater \textbf{0.8} &
  $\textbf{54.24} \pm \textbf{2.23}$ \\
 &  &  &  &                           & \textgreater 0.9 & $48.0 \pm 2.10$  \\ \cline{5-7} 
 &  &  &  & \multirow{2}{*}{VGG19}    & \textgreater 0.8 & $42.36 \pm 1.02$   \\
 &  &  &  &                           & \textgreater 0.9 & $41.07 \pm 5.21$  \\ \cline{5-7} 
 &  &  &  & \multirow{2}{*}{ResNet50} & \textgreater 0.9 & $28.42 \pm 4.84$  \\
 &  &  &  &                           & \textgreater 0.8 & $30.23  \pm 3.0$ \\ \hline


\end{tabular}
\caption{Counter Experiment 02: Faster RCNN}
\label{ce2}
\end{table}

\subsection{Prime Experiment: Dual Phase Approach}\label{prime_experiment}
Prime experiment has been performed by creating a pipeline in two phases shown in Figure. \ref{fig: FasterRCNN.pdf}. In the first phase, segmentations of grain has been performed and the segmented parts were cropped and saved as the secondary dataset, mentioned in subsubsection \ref{DDFG}. K fold cross-validation (K=5) have been performed where train and validation split was 80:20. As a result, the full primary dataset has been converted into a secondary dataset. In the second phase, three selected CNN architectures have been utilized for final classification after labelling the secondary dataset in terms of three classes. 

\begin{figure}
\captionsetup[subfigure]{labelformat=empty}
 \centering
 \begin{subfigure}{1.0\textwidth}
  \centering
  \includegraphics[width=1.0\linewidth]{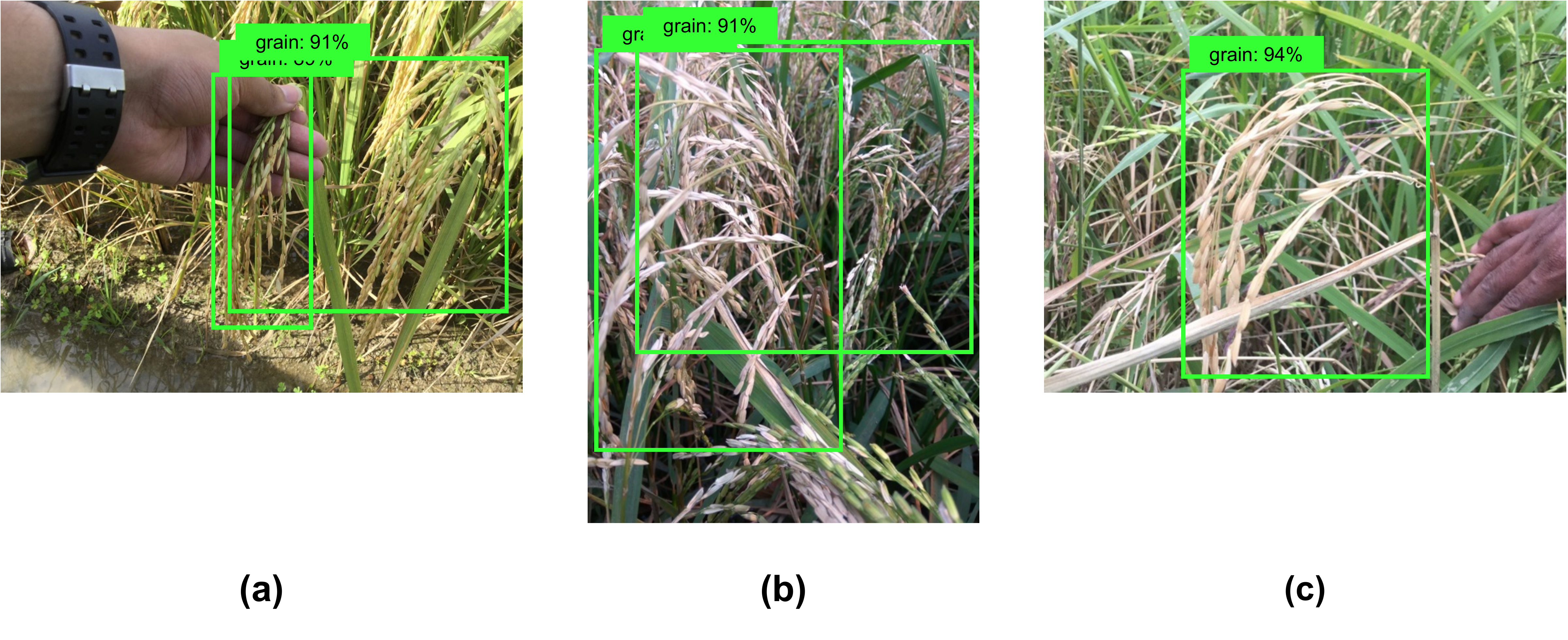}
 \end{subfigure}
\caption{Prime Experiment: First Phase; Sample Outcome: accuracy of detected objects. Phase One detected two bounding boxes from (a) and (b) as the second bounding box meets IoU and accuracy threshold. (c) have not met the IoU or accuracy threshold so the second bounding box is absent.}\label{frcnn_error}
\end{figure}

\subsubsection{Phase One: Segmentation of Grains}\label{methods_ph1}
The goal of phase one is to crop out the significant part (grain) from a particular image.
Faster RCNN have been utilized following three different CNN architecture as the backbone.
Applied CNN architectures were VGG16, VGG19 and ResNet50 which were already applied in counter experiment 02 mentioned in Subsubsection \ref{counter02}. Also, hyperparameters setting have been followed from the counter experiment 02. Only the best performed hyperparameter setting were applied in phase one mentioned in Table \ref{table_hyperparam}. Faster RCNN with VGG16 as backbone achieved the best mAP score of $84.3 \pm 2.36$. The result have been achieved through five fold cross validation. Thus all images in the dataset have been evaluated. From each image at most two new images have been generated which creates a secondary dataset. This operation has been performed by selecting two best bounding boxes from each image. First bounding box is the best bounding box referred by the Faster RCNN which is cropped and became the part of the new dataset. Second bounding box has been selected which satisfy the IoU threshold of 0.5 and accuracy threshold of 90\%. For several images there were no bounding boxes which met this requirement. On that case only one image have been selected for the new dataset.  
Figure. \ref{frcnn_error} shows the bounding boxes from each image for phase one. Here on sub figure (c) only one bounding box get detected.

\begin{table}
\centering
\begin{tabular}{llllll}
\hline
\begin{tabular}[c]{@{}l@{}}Anchor\\ Box Ratio\end{tabular} &
  \begin{tabular}[c]{@{}l@{}}Anchor \\ Box Pixels\end{tabular} &
  \begin{tabular}[c]{@{}l@{}}RPN \\ Threshold\end{tabular} &
  \begin{tabular}[c]{@{}l@{}}CNN\\ Architecture\end{tabular} &
  \begin{tabular}[c]{@{}l@{}}Overlap\\ Threshold\end{tabular} &
  \begin{tabular}[c]{@{}l@{}}mAP (\%)\end{tabular} \\ \hline
\multirow{3}{*}{\begin{tabular}[c]{@{}l@{}}(1:1),(\begin{math}\frac{1}{\sqrt2}:\frac{2}{\sqrt2}\end{math}), \\(\begin{math}\frac{2}{\sqrt2}:\frac{1}{\sqrt2}\end{math}),(2:2))\end{tabular}} &
  \multirow{3}{*}{\begin{tabular}[c]{@{}l@{}}32, 64,\\ 128, 256\end{tabular}} &
  \multirow{3}{*}{0.4 - 0.8} &
  \textbf{VGG16} &
  \textgreater \textbf{0.8} &
  $\textbf{84.3} \pm \textbf{2.36}$ \\
 &  &  & VGG19    & \textgreater 0.8 & $73.5 \pm 1.34$ \\
 &  &  & ResNet50 & \textgreater 0.8 & $65.9 \pm 2.59$ \\ \hline
\end{tabular}
\caption{Prime Experiment: Phase One}
\label{p1}
\end{table}

\subsubsection{Phase Two: Classification}
Image data received from phase one channeled through phase two where it provides the classification result.
Again three different CNN architecture, VGG16, VGG19 and ResNet50 have been applied in this phase.
Best settings from counter experiment 01 have been reapplied in this phase that has been mentioned in Subsubsection  \ref{counter01}. VGG16 emerged with the best validation accuracy of  $88.11 \pm 3.86$ mentioned in Table \ref{p2}. 

\begin{table}
\centering
\begin{tabular}{lllll}
\hline
\begin{tabular}[c]{@{}l@{}}CNN\\ Architecture\end{tabular} &
  \begin{tabular}[c]{@{}l@{}}Train \\ Loss\end{tabular} &
  \begin{tabular}[c]{@{}l@{}}Train\\ Accuracy \\(\%)\end{tabular} &
  \begin{tabular}[c]{@{}l@{}}Validation\\ Loss\end{tabular} &
  \begin{tabular}[c]{@{}l@{}}Validation\\ Accuracy \\(\%)\end{tabular} \\ \hline
\textbf{VGG16}    & \textbf{0.196} & \textbf{94.47} & \textbf{0.195} & $\textbf{88.11} \pm \textbf{3.86}$ \\
VGG19    & 0.095 & 89.98 & 0.093 & $86.43 \pm 2.98$ \\
ResNet50 & 0.367 & 89.63 & 0.281 & $78.00 \pm 2.32$ \\ \hline
\end{tabular}
\caption{Prime Experiment: Phase Two}
\label{p2}
\end{table}

\begin{figure}
\captionsetup[subfigure]{labelformat=empty}
 \centering
 \begin{subfigure}{1.0\textwidth}
  \centering
  \includegraphics[width=1.0\linewidth]{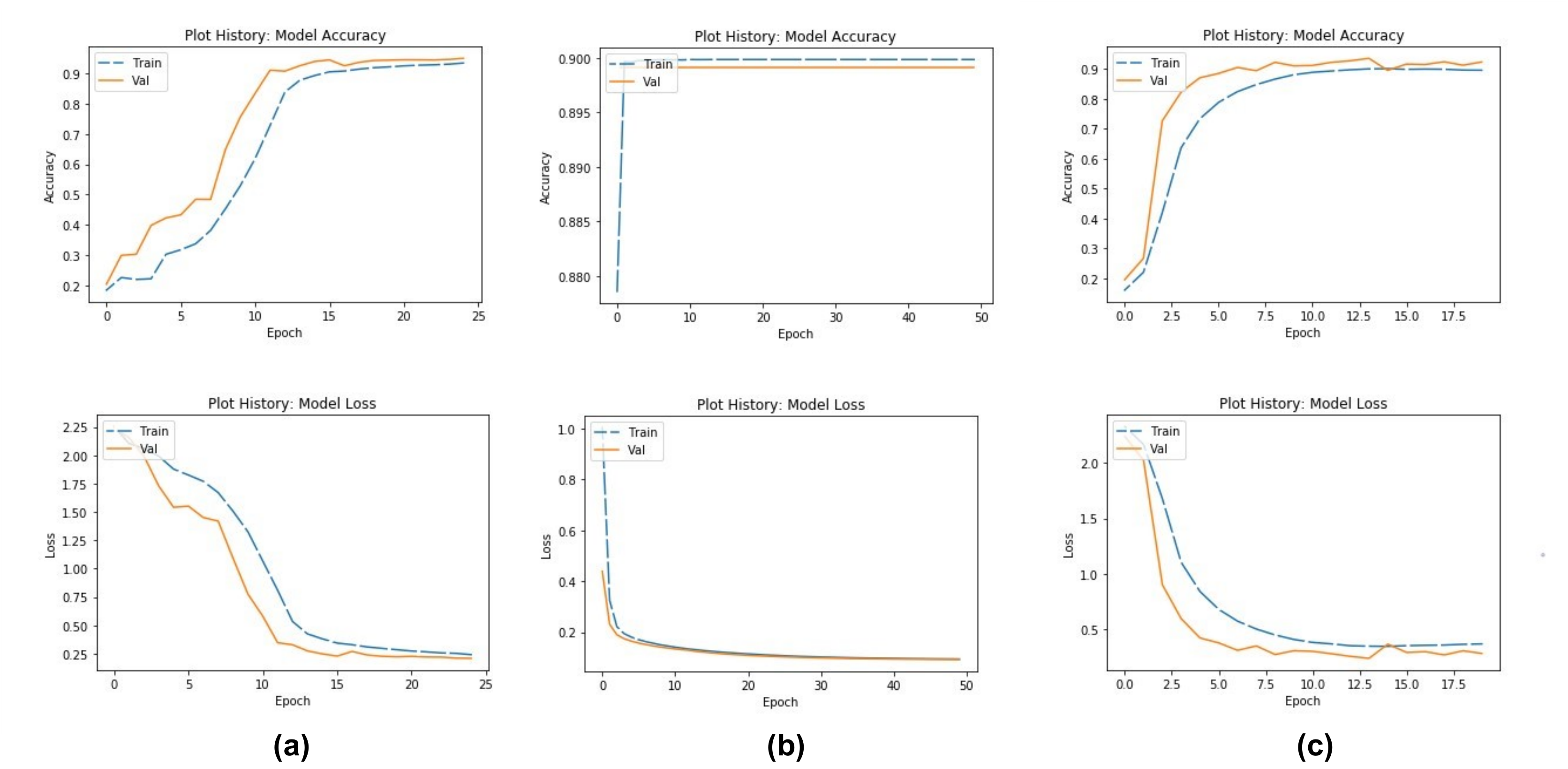}
 \end{subfigure}
\caption{Accuracy and loss for train and validation; Phase Two.(a)VGG16, (b)VGG19, (c)ResNet50.}\label{fig: loss_main.pdf}
\end{figure}

\begin{figure}
\captionsetup[subfigure]{labelformat=empty}
 \centering
 \begin{subfigure}{1.0\textwidth}
  \centering
  \includegraphics[width=1.0\linewidth]{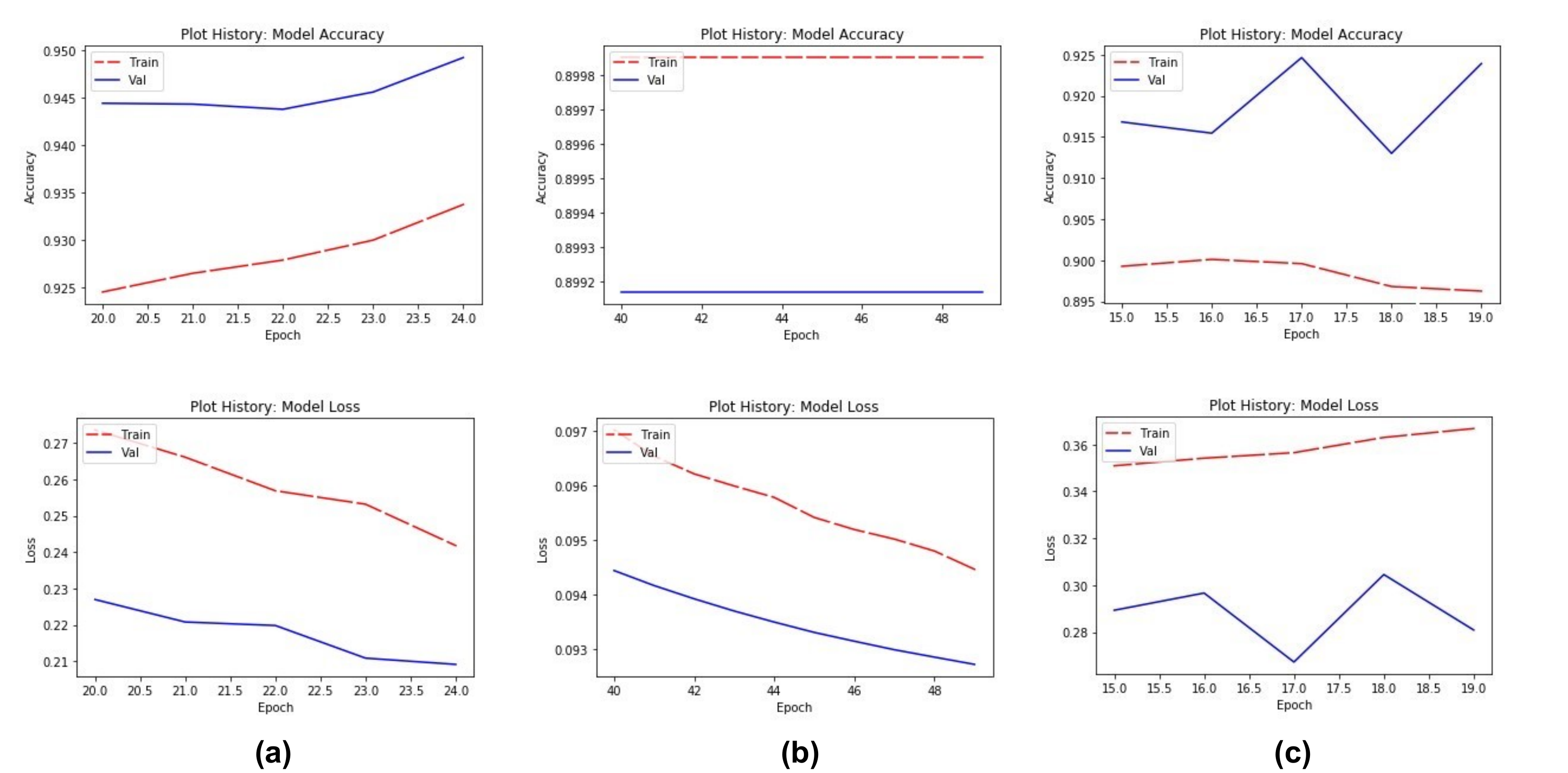}
 \end{subfigure}
\caption{Accuracy and loss for train and validation (Zoomed); Phase Two. Only VGG 16 is showing further improvement is possible through training.(a)VGG16, (b)VGG19, (c)ResNet50.}\label{fig: loss_slow.pdf}
\end{figure}

Figure. \ref{fig: loss_main.pdf} shows loss and accuracy graph for train and validation of the first training fold out of five folds for phase two. The graphs also shows the training was less time consuming as the dataset was small. By zooming in the graph it is visible that VGG16 was still learning shown in Figure. \ref{fig: loss_slow.pdf}. 

In general, expectancy from CNN models like VGG16 is higher in this dataset as there is only 3 class and they are quite different from each other in terms of class features. This is a pipeline-based process and phase one can generate FP/FN results which will be channeled through phase two. As a result, phase two will be unable not classify them properly. Which is a limitation of this system. This issue can be solved by presenting a new class which can be titled "No Grain" that will declare anything but grain.

\section{Conclusion}\label{conclusion}

In brief, this research has the following contributions:

– A dual phase approach capable of learning from small rice grain disease dataset has been proposed.

– A smart segmentation procedure has been proposed in phase one which is capable of handling heterogeneous background prevalent in plant disease image dataset collected in real life scenario

– Experimental comparison has been provided with straightforward use of state-of-the-art CNN architectures on the small rice grain dataset to show the effectiveness of the proposed approach.

\section{Acknowledgments}
We thank Information and Communications Technology (ICT) division, Bangladesh for aiding this research and the authority of Bangladesh Rice Research Institute (BRRI) for supporting us with field level data collection. We also acknowledge the help of RMIT University who gave us the opportunity to use their GPU server.

\bibliographystyle{unsrt}  
\bibliography{references}  

\end{document}